\documentclass{article}
\usepackage{ijcai25}

\usepackage{times}
\usepackage{soul}
\usepackage{url}
\usepackage[hidelinks]{hyperref}
\usepackage[utf8]{inputenc}
\usepackage[small]{caption}
\usepackage{graphicx}
\usepackage{amsmath}
\usepackage{amssymb}
\usepackage{amsthm}
\usepackage{booktabs}
\usepackage{algorithm}
\usepackage{algorithmic}
\usepackage{multirow}
\usepackage[switch]{lineno}
\usepackage{makecell}
\usepackage{appendix}
\usepackage{svg}
\renewcommand{\arraystretch}{1.4}

\title{Stable CDE Autoencoders with Acuity Regularization for Offline Reinforcement Learning in Sepsis Treatment}

\author{
    Yue Gao
    \affiliations
    Keebo.AI
    \emails
    gao12@ualberta.ca
}

\begin{document}

\maketitle

\begin{abstract}
Effective reinforcement learning (RL) for sepsis treatment depends on learning stable, clinically meaningful state representations from irregular ICU time series. While previous works have explored representation learning for this task, the critical challenge of training instability in sequential representations and its detrimental impact on policy performance has been overlooked. This work demonstrates that Controlled Differential Equations (CDE) state representation can achieve strong RL policies when two key factors are met: (1) ensuring training stability through early stopping or stabilization methods, and (2) enforcing acuity-aware representations by correlation regularization with clinical scores (SOFA, SAPS-II, OASIS). Experiments on the MIMIC-III sepsis cohort reveal that stable CDE autoencoder produces representations strongly correlated with acuity scores and enables RL policies with superior performance (WIS return $> 0.9$). In contrast, unstable CDE representation leads to degraded representations and policy failure (WIS return $\sim$ 0). Visualizations of the latent space show that stable CDEs not only separate survivor and non-survivor trajectories but also reveal clear acuity score gradients, whereas unstable training fails to capture either pattern. These findings highlight practical guidelines for using CDEs to encode irregular medical time series in clinical RL, emphasizing the need for training stability in sequential representation learning.
\end{abstract}

\section{Introduction}

Sequential decision-making is a cornerstone of modern healthcare, particularly in dynamic clinical scenarios, where timely diagnosis and adaptive treatment strategies are essential for patient survival \cite{Shashikumar2021,Solis2023}. A representative case is sepsis management, which requires timely diagnosis and appropriate treatment strategies. Reinforcement learning (RL) offers a promising framework for modeling sequential decision-making in clinical settings, where treatment strategies must adapt over time to a patient's evolving condition. In the context of sepsis management, RL has been employed to derive policies that optimize interventions including fluid resuscitation and vasopressor administration. For instance, the \textit{AI Clinician} \cite{Komorowski2018Nature}, demonstrated RL’s potential by training on the Medical Information Mart for Intensive Care III (MIMIC-III) dataset to recommend treatment strategies that outperformed clinician baselines. Subsequent advances in deep and distributional RL further refined policy robustness and flexibility, underscoring RL’s role in sepsis management \cite{killian20a,Jayaraman2024RL,Bock2022}. These advancements underscore the potential of RL to generate effective sequential treatment decisions in critical care environments. 

In this work, we use MIMIC-III \cite{Johnson2016MIMICIII,Johnson2016ScientificData,Goldberger2000PhysioNet} as experimental data, which containts clinical data from 40,000+ ICU patients, with vital signs, lab results, and treatments recorded at irregular time intervals. It provides a rich foundation for developing data-driven sepsis management tool. However, a key challenge lies in constructing informative state representations from this noisy, irregularly sampled data. Prior work has explored recurrent neural networks (RNNs) and autoencoders for this task, but their instability during training often leads to suboptimal representations and degraded policy performance \cite{Bock2022,Solis2023,killian20a}.

Neural Controlled Differential Equations (Neural CDEs) are state-of-the-art models for irregular time series data due to their ability to model continuous-time dynamics and handle irregular sampling \cite{killian20a,morrill2021}. Unlike discrete-step RNNs, CDEs use differential equations to propagate hidden states, capturing latent physiological trends more accurately. This makes them particularly suited for sepsis management, where patient states evolve smoothly between sparse observations. However, CDEs are prone to training instability such as gradient explosion or collapse when unregularized or trained for excessive epochs \cite{Susama2022}. This instability stems fundamentally from numerical challenges in solving the underlying differential equations \cite{SHOOSMITH2003}, particularly when backpropagating through adaptive ODE solvers \cite{baker2022proximalimplicitodesolvers}. 

\begin{figure}[t]
    \centering
    \includegraphics[width=\columnwidth]{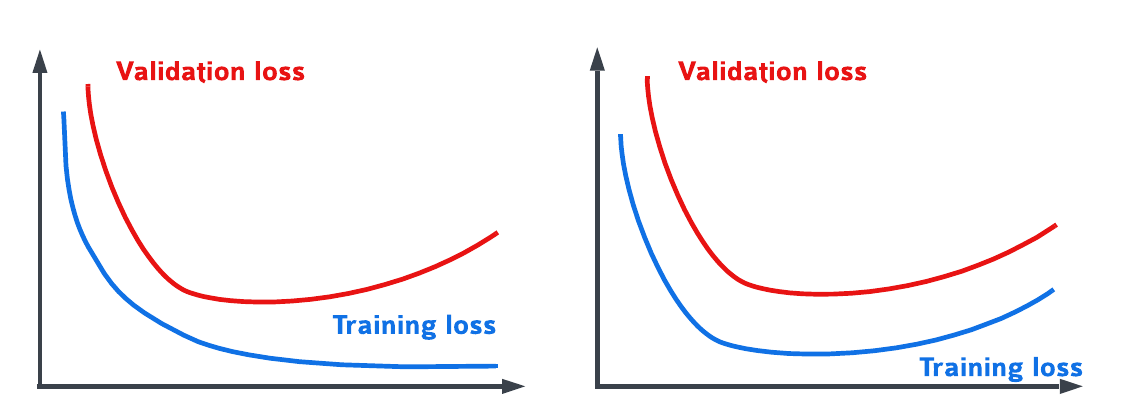}
    \caption{Distinguishing training failure modes: (Left) Classical overfitting; (Right) Numerical instability.}
    \label{fig:overfitting_vs_instability}
\end{figure}
\begin{figure*}[t]
    \centering
    \includegraphics[width=\textwidth]{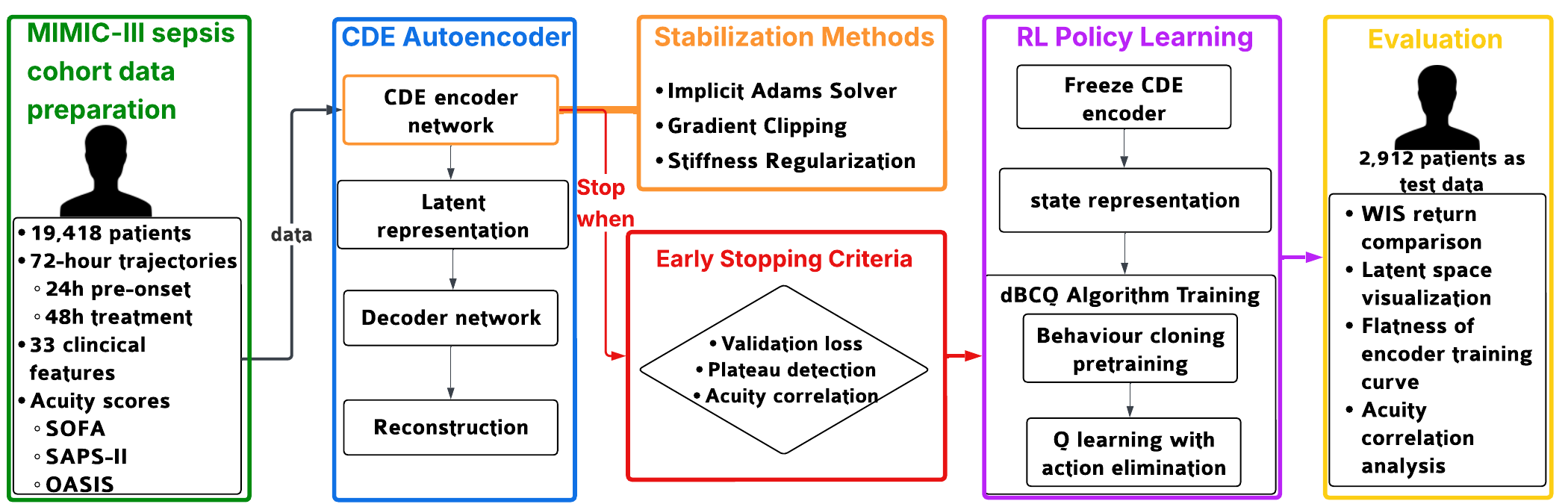}
    \caption{Overall architecture of the proposed framework.}
    \label{fig:workflow}
\end{figure*}
A key challenge in applying Neural CDEs to clinical time series is distinguishing between numerical instability and model overfitting, which are two fundamentally different failure modes. Overfitting appears when the training loss continues to decrease while the validation loss starts going upward; In contrast, CDEs exhibit unique instability patterns where both losses increase simultaneously (Figure~\ref{fig:overfitting_vs_instability}). This instability occurs when numerical solvers fail to handle stiffness in the learned vector fields \cite{SHOOSMITH2003}.
For clinical time series modeling, this instability is particularly crucial, as trajectory smoothness and sudden acuity transitions must both be captured. 

This limitation in CDE state representation was overlooked in prior studies. For instance, \cite{killian20a} compared CDEs against other autoencoders for sepsis state representation and showed its superiority, but did not address their instability, leading to inconsistent results and an erroneous dismissal of clinical acuity score regularization.

In this work\footnote{Code available at \url{https://github.com/GAOYUEtianc/RL_mimic_CDE_stable}}, we use the MIMIC-III database to train, evaluate, and compare CDE representations for sequential state encoding, with the goal of improving discrete Batch-Constrained Q-learning (dBCQ) models for offline sepsis treatment decision-making, with the following key contributions:
\begin{enumerate}
    \item Identify and address the critical challenge in clinical RL: training instability in Neural CDEs for time-series state representation. Proposed a rigorous early stopping strategy that reliably produces robust representations.
    \item Demonstrate that clinical acuity scores (SOFA/SAPS-II/OASIS) effectively regularize CDE autoencoder training when properly stabilized, contradicting prior claims about their ineffectiveness in RL representations.
    \item Establish the consistency in MSE loss trend and acuity correlation loss trend during CDE autoencoder training process, validating acuity scores as meaningful proxies for patient state severity in CDE state representations.
    \item Compare three stabilization methods (gradient clipping, implicit solvers, stiffness regularization) for clinical time series representation training, and provide practical guidance for selecting methods based on impact on downstream RL policy performance.
\end{enumerate}
Our findings revisit and clarify inconsistencies reported in prior works such as \cite{killian20a}, which concluded that acuity-based regularization fails to improve CDE representations. We show this claim likely arose from unstable training regimes that obscure representation quality. By addressing CDE instability, we show that acuity regularization improves both state representation quality and RL performance.
\section{Background}
\subsection{MIMIC‑III Data Irregular Properties}
The MIMIC‑III repository contains ICU patient data including physiological signals, laboratory results, medications, and interventions as multivariate time series \cite{Johnson2016MIMICIII}. However, these measurements were recorded at irregular intervals based on clinical needs, resulting in uneven sampling patterns and substantial missing data \cite{Johnson2016ScientificData,NEURIPS2019_ODE}.
The dataset contains inherent measurement noise and informative missing patterns, where missing observations could provide clinical insights. This creates a low signal-to-noise ratio that makes direct trajectory modeling challenging \cite{Che2018GRUD,Yoon2018GAIN}. Different variables are recorded at varying frequencies, causing covariate shift and temporal aliasing that reduce model generalization \cite{Zhang2019FeatureInteraction}. Simple imputation techniques like forward‑filling or mean substitution fail to capture complex temporal dependencies and often introduce bias \cite{Yoon2018GAIN}. 

To mitigate these challenges, researchers have developed several approaches. Time‑aware recurrent models use masking and time interval embeddings to explicitly handle missingness and irregularity \cite{Che2018GRUD}. Continuous‑time methods such as ODE‑RNNs and latent ODEs learn differential equations that naturally adapt to varying time intervals \cite{NEURIPS2019_ODE}. 
Among these, Neural Controlled Differential Equations offer a SOTA approach to encode MIMIC‑III’s high signal-to-noise trajectories. They produce well-representated states suited for downstream offline reinforcement learning in sepsis management \cite{NEURIPS2019_ODE,kidger21b,killian20a}.

\subsection{Neural Controlled Differential Equations (Neural CDEs) as State Encoders}\label{sec:CDE}
Neural Controlled Differential Equations are a class of continuous-time models that generalize recurrent neural networks by defining hidden state dynamics through differential equations \cite{kidger21b}.
We initialize the hidden state as $h(t_{0}) = Wo(t_{0})+b$, using a learnable linear map $g:\mathbb{R}^{d}\rightarrow \mathbb{R}^{h}$ implemented as a single-layer neural network.
Then for the continuous-time irregular observations $o_t\in \mathcal{O}$, CDE encodes a hidden state $h(t)\rightarrow \mathcal{H}$ in differential form
via\begin{align}
    \partial h(t) = f_{\theta}(h(t))\partial o_t
\label{eq:CDEformula}
\end{align}
where $f_{\theta}$ is a neural network parameterizing the system dynamics, and the differential $\partial o_t$ accounts for irregular sampling intervals \cite{NEURIPS2018_ODE,NEURIPS2019_ODE}. 
The CDE acts as an encoder $\psi: \mathcal{O}\rightarrow \mathcal{H}$, mapping time-based observations to a latent space. A decoder $\phi : \mathcal{H} \rightarrow \mathcal{O}$ reconstructing subsequent observations\begin{align}
    \hat{o}_{t} = \phi(h(t))
\label{eq:reconstruct_observation}
\end{align}
is trained to minimize the loss $\mathcal{L}_{\text{MSE}}(o_{t},\hat{o}_{t}) = || o_{t} - \hat{o}_{t}||^2$. This autoencoding framework ensures $h(t)$ retains clinically relevant information while discarding noise \cite{Lesort2018379}.
In this work, we use 4-th order Runge-Kutta (RK4) as the numerical solver for the CDE as a baseline \cite{Hairer2000RK4}.
\subsection{Instability in Neural CDE Training}\label{sec:CDE_instability_background}
During training, Neural CDEs can exhibit numerical instability due to the sensitivity of ODE solvers to the learned vector field $f_{\theta}$. If $f_{\theta}$ has large Lipschitz constant or long time sequences, the backpropagating gradients can explode, leading to erratic updates and divergence \cite{SHOOSMITH2003,Kim2021}.
Hence, picking a proper stabilization method that can balance the smoothness and sensitivity to rare high-acuity transitions is crucial.

Effective methodologies to stabilize Neural CDEs training include implicit solvers that handle stiffness by solving linearized equations at each step \cite{Solomon2021RK4ABM,baker2022proximalimplicitodesolvers,NEURIPS2018_ODE,fronk2024trainingstiffneuralordinary}, gradient clipping to prevent large updates \cite{NEURIPS2022Gradient}, and regularization techniques smoothing the learned vector field by penalizing high local error and stiffness \cite{pal2022openingblackboxacceleratingneural}.
Building on these insights, we employ early stopping and stabilization techniques to ensure Neural CDEs training produces high-quality state representations that generalize well and enable effective downstream RL performance.
\subsection{Reinforcement Learning with CDE State Representations}
Sequential sepsis treatment can be modeled as a Partially Observable Markov Decision Process (POMDP):
\begin{itemize}
    \item State : $s_t = h(t)$ (CDE encoded history of observations $o_{0:t}$).
    \item Action : $a_t\in \{1, ..., 25\}$ represents discrete combinations of intravenous fluids and vasopressor doses \cite{Komorowski2018Nature}
    \item Reward : \begin{equation}
        r_t = \begin{cases}
            +1 & \text{if patient survives at trajectory end}\\
            -1 & \text{if patient dies at trajectory end}\\
            0 & \text{otherwise (at intermediate steps)}
        \end{cases}
    \label{eq:reward_function}
    \end{equation}
    \item Policy: $\pi(a_t|s_t)$ maps states to probability distribution of actions, aiming to optimize cumulative reward via offline RL.
\end{itemize}

\subsection{Clinical Acuity Scores as Priors}\label{sec:clinical_score}
We leverage three established clinical acuity scores (SOFA, SAPS-II, and OASIS) as semi-supervision to regularize the CDE latent space. Detailed score definitions are provided in Appendix~\ref{app:acuity_score}.
We define the correlation loss:
\begin{align}
    \mathcal{L}_{\text{corr}}(\hat{s}_t) = -(\rho_{\text{SOFA}}(\hat{s}_t) + \rho_{\text{SAPS-II}}(\hat{s}_t) + \rho_{\text{OASIS}}(\hat{s}_t))
\label{eq:loss_corr_def}
\end{align}

\noindent The regularized total loss function is defined to be:
\begin{equation}
\begin{aligned}
    \mathcal{L}_{\text{total}}(o_{t}, \hat{o}_{t}) = \mathcal{L}_{\text{MSE}}(o_{t}, \hat{o}_{t}) + \lambda\cdot\mathcal{L}_{\text{corr}}(\hat{s}_t)
\label{eq:loss_func_with_acuity}
\end{aligned}
\end{equation}
where $\rho(\hat{s}_t)$ denotes the Pearson correlation between the latent state representation $\hat{s}_t$ and acuity score. For simplicity,  we denote the losses at epoch $i$ as $\mathcal{L}_{\text{MSE}}(i)$, $\mathcal{L}_{\text{corr}}(i)$, $\mathcal{L}_{\text{total}}(i)$.
\section{Methodology}
We present a framework as shown in Figure~\ref{fig:workflow} for sepsis treatment policy learning using stabilized CDE state representations, evaluated through offline RL and clinical interpretability metrics.
\subsection{Overall Architecture}
Our framework builds on prior work in sepsis treatment RL \cite{killian20a,Bock2022,Jayaraman2024RL} but addresses a critical gap: the training instability of Neural CDEs for state representation. While existing approaches use CDEs to encode patient history into hidden states and decode future observations, they overlook how training instability affects both representation quality and downstream policy performance.
We identify two key instability symptoms:
\begin{itemize}
\item Unpredictable fluctuations in observation prediction loss
\item Weakened correlation between learned states and clinical acuity scores
\end{itemize}
To address these, we introduce:
\begin{itemize}
\item Training stabilization via early stopping and specialized techniques
\item Acuity-aware regularization to maintain clinical relevance
\end{itemize}
The stabilized representations then feed into Batch Constrained Q-learning, where we evaluate their impact on final policy performance through both quantitative metrics and clinical interpretability measures.
\subsection{Data Preparation}
Our study utilizes the MIMIC-III v1.4 critical care database, processed according to the established sepsis cohort definition from prior work \cite{Komorowski2018Nature}. After processing, the dataset comprises 19,418 adult sepsis patients, with each patient trajectory spanning a clinically relevant 72-hour window around sepsis onset : capturing 24 hours preceding identification through 48 hours of subsequent treatment. These trajectories reflect the real-world challenges of ICU care, exhibiting irregular sampling intervals and heterogeneous measurements across 33 time-varying physiological features. The action space follows prior work in discretizing clinical interventions into 25 distinct combinations of intravenous fluids and vasopressor doses, binned by percentile ranges to maintain clinically meaningful groupings while enabling reinforcement learning. Patient outcomes define trajectory termination, with mortality recorded for the 9.2\% of patients who died within 48 hours of their final observation, while survivors comprise the remaining 90.8\%. 
To ensure faithful evaluation while preserving outcome distributions, we split the cohort into training (70\%, n=13,593), validation (15\%, n=2,913), and test sets (15\%, n=2,912), maintaining identical 9.2\% mortality rates across splits.
Clinical acuity scores (SOFA, SAPS-II, OASIS) for illness severity are calculated at each timestep using validated implementations that transform the 33 raw features into standardized risk metrics.
 \subsection{CDE Autoencoder Training with Rigorous Stopping Criteria}\label{sec:cde_stopping_criteria}
 The CDE autoencoder is trained to learn clinically meaningful state representations through a carefully designed optimization process that addresses the inherent instability of continuous-time neural networks. \\
\textbf{Continuous-Time Encoding Process} Our CDE autoencoder learns continuous-time state representations through a neural controlled differential equation (Equation~\ref{eq:CDEformula}), implemented as a 3-layer network with ReLU activations and layer normalization. The encoder outputs a hidden state $h(t) \in \mathbb{R}^d$ where $d$ is the representation dimension.\\
\textbf{Early Stopping Strategy} To prevent training instability while preserving signal capture, we propose a multi-criteria early stopping method that selects the optimal checkpoint epoch $e^*$ when all conditions are first met:
\begin{enumerate}
  \item \textbf{Near-optimal validation loss:} 
  \begin{align}\mathcal{L}_{\text{val}}(e^*) \leq \min_{e\leq e^{*}}\mathcal{L}_{\text{val}}(e) + \epsilon_1
   \end{align}
  \item \textbf{Stable training plateau:} For the last $p$ epochs, the total loss variation remains within $\epsilon_2$ fraction of the minimum loss:
  \begin{align}
    \left| 
        \max_{e^* - p \leq i \leq e^*} \mathcal{L}_{\text{total}}(i)
        - \min_{e^* - p \leq i \leq e^*} \mathcal{L}_{\text{total}}(i)
    \right|\leq&  \nonumber \\
    \epsilon_2 \cdot
        \min_{e^* - p \leq i \leq e^*} \mathcal{L}_{\text{total}}(i)&
\end{align}
  \item \textbf{Clinically meaningful representations over training:} Mean acuity score correlation on train set exceeds a threshold. 
  \begin{align}
    \rho(e^{*}) \geq \rho_{\text{threshold}}
  \end{align}
\end{enumerate}
\noindent We train CDE autoencoders across multiple random seeds through comprehensive hyperparameter tuning (learning rates, hidden sizes, $\epsilon_{1},\; \epsilon_{2},\; p, \;\rho_{\text{threshold}}$), then select the best performing configuration evaluated via the ultimate RL policy measured by WIS return \footnote{Weighted Importance Sampling (WIS) is a technique used in offline reinforcement learning to estimate the expected return of a target policy using data collected from a different behavior policy. By normalizing importance weights across trajectories, WIS reduces variance compared to ordinary importance sampling, albeit introducing some bias. This trade-off often results in more stable and reliable policy evaluations \cite{NIPS_WIS2014}.}.
\subsection{Stabilization Methods for CDE Training}\label{sec:stabilizing_method}
To improve training stability and facilitate model selection, we apply stabilization techniques to the CDE encoder, specifically on the vector field $f_{\theta}$ that governs hidden state dynamics. We implement three stabilization techniques:
\begin{enumerate}
\item \textbf{Gradient Clipping}: Constrains extreme gradient updates to prevent sudden spikes during backpropagation
\item \textbf{Implicit Adams Solver}: Uses an implicit numerical integration scheme to handle stiffness in the continuous-time dynamics
\item \textbf{Stiffness Regularization}: Directly penalizes high curvature in the learned vector field
\end{enumerate}
Appendix~\ref{app:stabilization_methods} introduces those methodologies in detail.

\noindent\textbf{Evaluation Protocol:} We conduct comprehensive hyperparameter tuning for hyperparameters in Table~\ref{tab:stabilization_hparams} across \begin{itemize}
    \item Multiple random seeds : $25, 53, 1234, 2020$
    \item Learning rates : ${1\times10^{-4}, 2\times10^{-4}, 5\times10^{-4}}$
    \item Hidden dimensions : ${4, 8, 16, 32, 64, 128}$
    \end{itemize}
Each configuration undergoes 100 epochs training until reaching early stopping criteria in Section~\ref{sec:cde_stopping_criteria}, and is evaluated using two stability metrics:
\begin{enumerate}
\item \textbf{Plateau Length}: Number of consecutive epochs where the training loss remains on a plateau:
\begin{align}
| \mathcal{L}_{\text{total}}(i) - \min\mathcal{L}_{\text{total}} | \leq \epsilon_{2} \cdot \min\mathcal{L}_{\text{total}}
\label{eq:plateau_length}
\end{align}

\item \textbf{Mean Absolute Slope on Plateau}:
\begin{align}
S_{1} = \frac{1}{T} \sum_{i}^{T} | \Delta \mathcal{L}_{\text{total}}(i) |
\label{eq:MAS}
\end{align}
where $\Delta \mathcal{L}_{\text{total}}(i) = \mathcal{L}_{\text{total}}(i+1) - \mathcal{L}_{\text{total}}(i)$ and $T$ is the plateau length.
\end{enumerate}

\noindent The final model evaluation considers both these stability metrics and the downstream RL policy performance (WIS return).
\begin{table}[t]
\centering
\begin{tabular}{p{2.5cm} p{2.3cm} p{2.6cm}}
\toprule
\textbf{Method} & \textbf{Hyperparameter} & \textbf{Search Range} \\
\midrule
Gradient clipping & \textit{max norm }: $\tau$ & $\{0.1, 0.5, 1.0, 1.5\}$ \\
\makecell[l]{Implicit Adams\\ solver} & \textit{step size} : $\Delta t$ & $\{1/8, 1/4, 1/2\}$ \\
Stiffness regularization & $\lambda_{\text{reg}}$ & $\{0.005, 0.01, 0.015\}$ \\
\bottomrule
\end{tabular}
\caption{Stabilization methods and hyperparameters to be tuned.}
\label{tab:stabilization_hparams}
\end{table}

\subsection{dBCQ Policy Learning and Evaluation}
We employ the trained CDE encoder to transform raw patient trajectories into continuous state representations for offline reinforcement learning. Our policy learning approach builds on discrete Batch-Constrained Q-learning (dBCQ) \cite{killian20a}, which addresses key challenges in offline RL. The process is as follows:\\
\noindent \textbf{Behavior Cloning Pre-training}: We first train a policy to replicate the action distribution in the dataset, providing a conservative starting point.\\
\noindent \textbf{Constrained Q-Learning}: During policy optimization, the Q-function only considers actions that the behavior policy would likely take with probability $\geq \tau_{\text{BC}}$.\\
\noindent \textbf{Policy Evaluation}: We compute WIS returns of the trained Q-policy on the validation dataset to assess policy performance.
\section{Experiments and Empirical Results}
\subsection{Effect of CDE Early Stopping on Final RL Policy Result}\label{sec:exp_stopping_criteria}
We conducted extensive hyperparameter tuning to identify optimal training configurations. We performed a grid search over hidden state sizes $\{4,8, 16,32,64,128\}$, learning rates \(\{1\times10^{-4},2\times10^{-4},5\times10^{-4}\}\), acuity correlation coefficients $\lambda\in\{0, 0.5,1,1.5\}$, and stopping criteria parameters : $\epsilon_{1}\in \{0.05, 0.1, 0.15, 0.2\}, p\in \{20, 30, 40, 50\}, \epsilon_{2}\in \{0.02, 0.03, 0.04, 0.05\}, \rho_{\text{threshold}}\in \{0.6, 0.65, 0.7, 0.75\}$.

The best configuration (hidden size=64, learning rate=$2\times10^{-4}$, $\lambda=1$, $\epsilon_1=0.1$, $p=30$, $\epsilon_2=0.02$, $\rho_{\text{threshold}}=0.7$) achieved the highest mean WIS return. 
Using this configuration, we trained the CDE autoencoder for 100 epoches and recorded its $\mathcal{L}_{\text{total}}$ (Equation~\ref{eq:loss_func_with_acuity}) at each epoch.

\begin{figure}[t]
    \centering
    \includegraphics[width=\columnwidth]{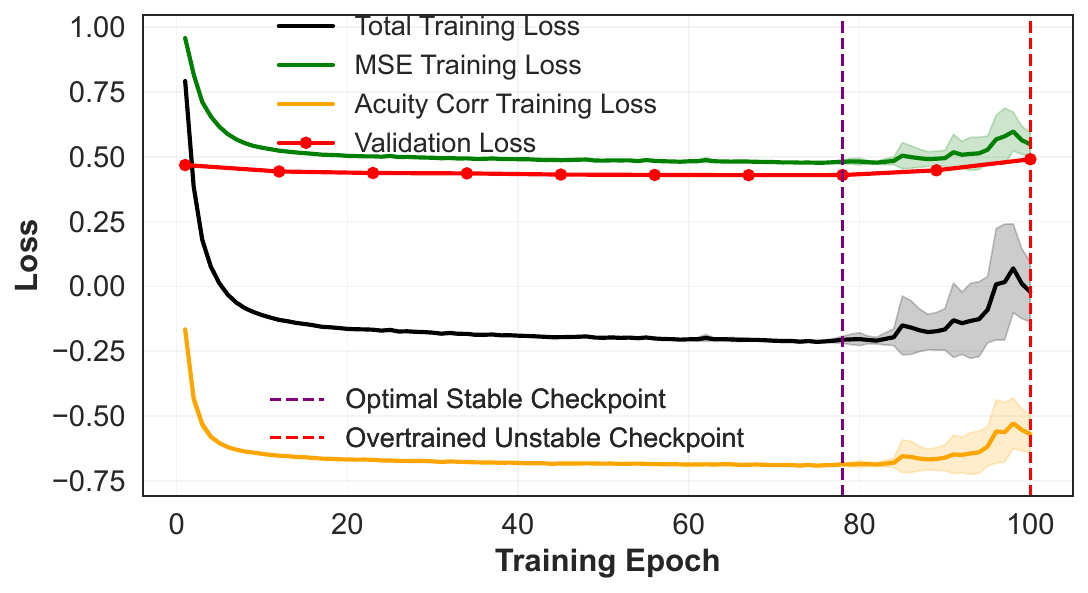}
    \caption{CDE autoencoder loss curves ($mean \pm std$  across multiple random seeds) versus training epoch with \textit{optimal and stable} \& \textit{overtrained and unstable} checkpoints.}
    \label{fig:cde_training_loss}
\end{figure}

Figure~\ref{fig:cde_training_loss} shows the training-epoch‐wise curves for the total loss, MSE loss, and correlation loss, with shaded bands indicating \(\pm1\) standard deviation across random seeds. To illustrate the impact of early stopping strategy, we highlight and compare those two checkpoints:  
\begin{enumerate}
    \item \textit{optimal and stable}: meets all stopping criteria in Section~\ref{sec:cde_stopping_criteria} with hyperparameters above, achieves a low, stable loss and high acuity correlation
    \item \textit{overtrained and unstable}: shows unstable loss patterns and degraded generalization
\end{enumerate}

\begin{figure}[t]
    \centering
    \includegraphics[width=\columnwidth]{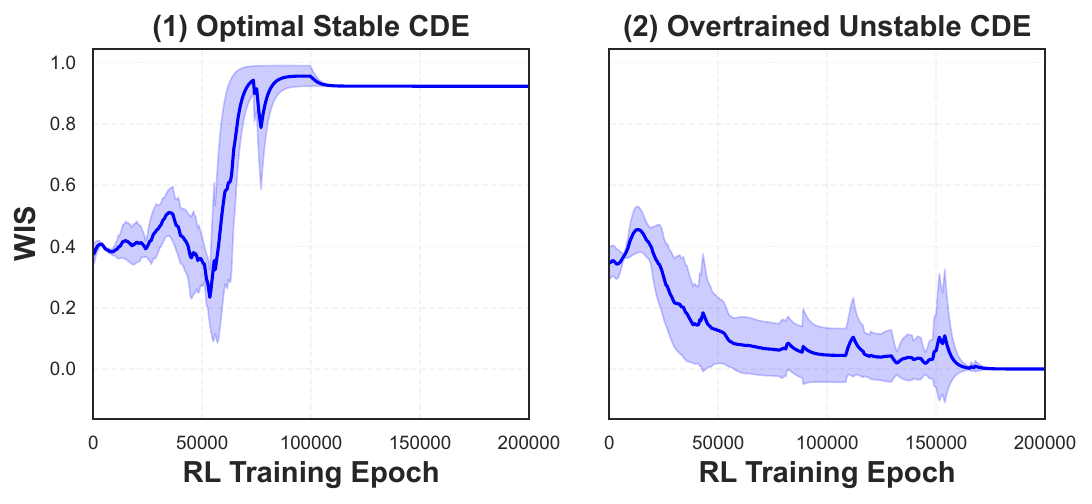}
    \caption{Evaluated WIS on validation set ($mean \pm std$ across random seeds) for dBCQ policies trained using \textit{optimal and stable} and \textit{overtrained and unstable} CDE state representations.}
    \label{fig:WIS_comparison}
\end{figure}

\noindent To evaluate how early stoppping affects downstream RL, we freeze each CDE encoder at the two checkpoints above and train a dBCQ RL policy for them respectively using identical setting: policy learning rate \(=1\times10^{-5}\), dBCQ action elimination threshold \(\tau_{\text{BC}}=0.3\), and epochs \(=2\times10^{5}\). Figure~\ref{fig:WIS_comparison} shows the WIS returns on validation patients trajectories versus RL training epoch, with shaded bands representing \(\pm1\) standard deviation across random seeds. The policy initialized from the \emph{optimal and stable} checkpoint consistently achieves high final return $0.9195$. In contrast, policies derived from the \emph{overtrained and unstable} CDE representations exhibits lower, more erratic RL performance, collapsing to a WIS return $3.2 \times 10^{-7}$, indicating complete value estimation failure.  
Collectively, these results demonstrate that selecting the right stopping epoch is essential for obtaining stable, generalizable representations from the CDE autoencoder, which has a critical impact on downstream RL policy quality.
And our proposed early stopping strategy (Section~\ref{sec:cde_stopping_criteria}) effectively optimized the state representation quality, yielding superior RL performance (WIS$\sim > 0.9$) compared to prior works \cite{killian20a,huang22apmlr} which reported WIS$\sim 0.775$.
\subsection{Clinical Alignment Through Early Stopping}
\label{sec:exp_acuity_alignment}
Building on our policy results, we analyze how early stopping affects the clinical interpretability of CDE representations. We compare latent spaces from \textit{optimal and stable} checkpoint and \textit{overtrained and unstable} checkpoint. All experiments in this section use the best hyperparameters from Section~\ref{sec:exp_stopping_criteria}, we use principal component analysis (PCA) to project those representations on validation trajectories (with first observation and end observation) into a lower dimensional space. 
\begin{figure}[t]
    \centering
    \includegraphics[width=\columnwidth]{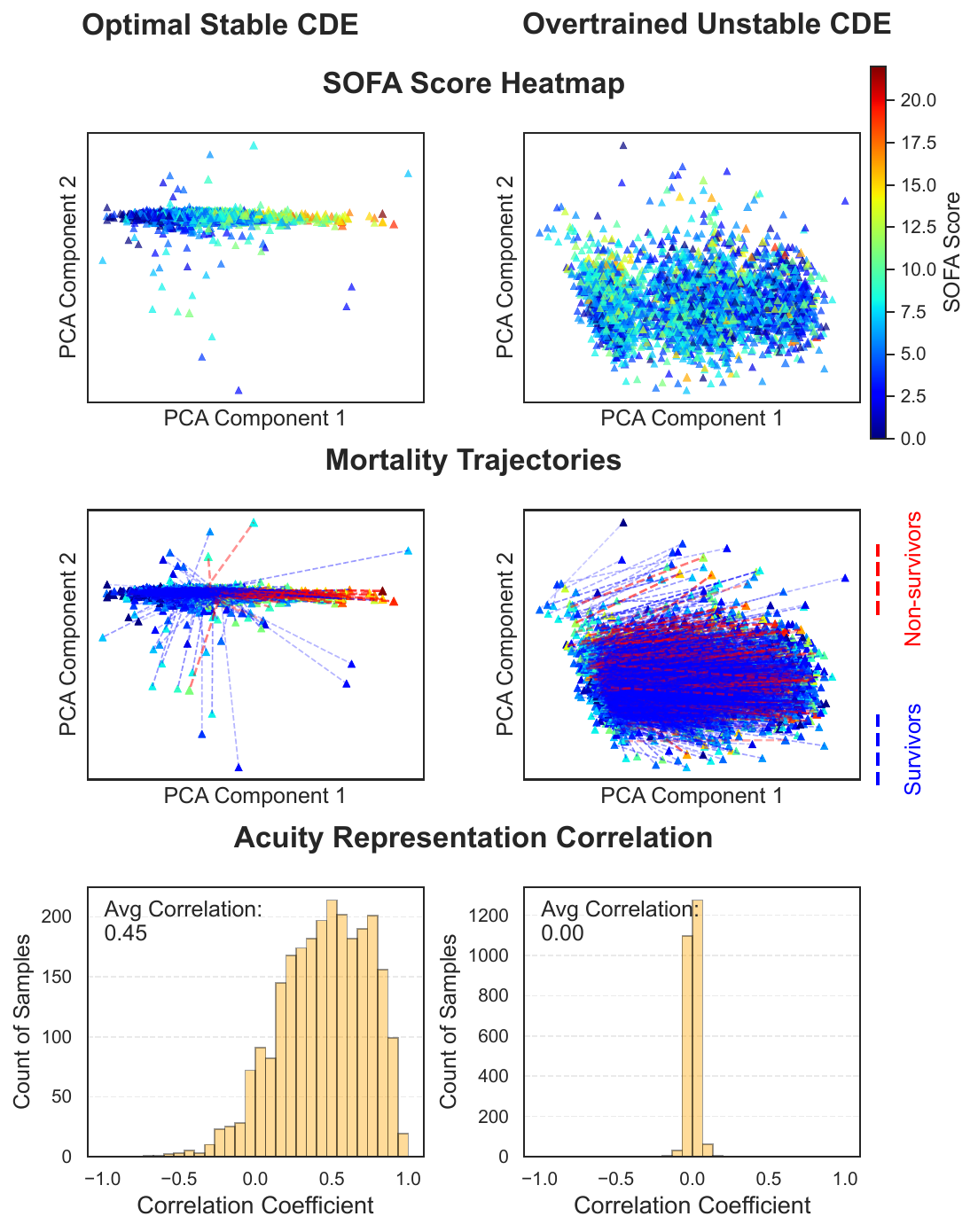}
    \caption{Comparative visualization of clinical alignment of \textit{optimal and stable} and \textit{overtrained and unstable} CDE representations on validation dataset trajectories}
    \label{fig:cde_pca_plot}
\end{figure}
Figure~\ref{fig:cde_pca_plot} shows three facets of the CDE representations, illustrates how stopping early affects clinical alignment on validation trajectories. In this section, we only show the result for SOFA score, but similar results hold for SAPS-II and OASIS as well (see Appendix~\ref{app:acuity_score_heatmap_app} for details).
\begin{enumerate}
\item\textbf{SOFA Score Distribution}: Each point is a validation observation colored by its SOFA score.  At the \textit{optimal and stable} CDE representation, patient states occupy a compact, well‐populated manifold, where samples with similar SOFA scores cluster together, revealing a smooth severity gradient. By contrast, the \textit{overtrained and unstable} CDE representation’s observations are sparse and unstructured, without clear pattern for SOFA score distribution.

\item\textbf{Mortality Trajectories}: We overlay each validation patient’s first‐to‐last latent jump, with survivors in blue and non‐survivors in red. In the \textit{optimal and stable} representation, non‐survivor paths cluster roughly in one zone, suggesting clear prognostic separation. In the \textit{overtrained and unstable} representation, both red and blue trajectories crisscross indiscriminately, offering no predictive signal.

\item\textbf{Acuity Representation Correlation}: We use histograms to show per‐validation-trajectory mean Pearson correlation between latent features and acuity scores. The \textit{optimal and stable} representation’s distribution skews to the right ($mean \; correlation \sim 0.45$). The \textit{overtrained and unstable} model’s distribution is centered around zero, with a tight unimodal peak, indicating that its latent features no longer capture any meaningful severity signal.
\end{enumerate}
In summary, early stopping strategy yields a CDE representation with better clinical interpretability. 

Also, this comparison indicates that acuity correlation serves as a reliable proxy for evaluating the clinical alignment of learned representations. The representation with clearer survivor separation and a structured SOFA score gradient also yields superior downstream RL policy performance. 
This is verified by Figure~\ref{fig:cde_training_loss} that the trend of MSE loss, acuity correlation loss, and validation loss are consistent. Quantitatively, we calculate the Pearson correlation coefficient and mutual information of $\mathcal{L}_{\text{MSE}}$ and $\mathcal{L}_{\text{corr}}$ over training epochs. As shown in Table~\ref{tab:mse_acuity_correlation}, the high Pearson correlation coefficient with significant p-value indicates a strong linear relationship; The mutual information of $1.62$ nats further supports this relationship, suggesting that $\mathcal{L}_{\text{MSE}}$ and $\mathcal{L}_{\text{corr}}$ are capturing the same underlying latent quality \cite{Young2023}. See Appendix~\ref{app:mse_acuity_correlation} for plot visualizing the correlation.
\begin{table}[t]
\centering
\begin{tabular}{ccc}
\toprule
\makecell[c]{\textbf{Pearson corr.}\\\textbf{coefficient r}} 
& \makecell[c]{\textbf{significant}\\\textbf{ p-value}}  & \textbf{mutual information} \\
\midrule
$0.9578\pm 0.0127$ & $<10^{-45}$ & $1.5296\pm 0.1847$ nats\\
\bottomrule
\end{tabular}
\caption{Correlation between MSE training loss $\mathcal{L}_{\text{MSE}}$ and acuity correlation training loss $\mathcal{L}_{\text{corr}}$ during CDE autoencoder training, averaged across random seeds.}
\label{tab:mse_acuity_correlation}
\end{table}

\begin{table}[t]
    \centering
    \begin{tabular}{p{2.1cm} p{2.3cm} p{2.3cm}}
        \toprule
        & \textbf{All Epochs} & \textbf{Plateau Epochs} \\
        \midrule
        Pearson Corr. & $0.6241 \pm 0.1934$ & $0.9284 \pm 0.0295$ \\
        p-value & $<0.05$ & $<0.05$ \\
        \bottomrule
        \end{tabular}
    \caption{Correlation between training loss $\mathcal{L}_{\text{total}}$ and validation loss $\mathcal{L}_{\text{val}}$ over CDE autoencoder training epochs, averaged across random seeds, compared on plateau V.S. whole training epochs.}
    \label{tab:training_validation_correlation}
    \end{table}
\noindent\textbf{Preventing overfitting.} As shown in Table~\ref{tab:training_validation_correlation}, training-validation loss correlation improves from moderate ($\sim0.62$) on global to near-perfect ($\sim 0.93$) on the plateau. This demonstrates our stopping criteria not only prevent instability but also avoid overfitting and ensure reliable generalization.

\noindent\textbf{Contradicting prior work.} Notably, this observation runs counter to claims in prior work \cite{killian20a}, which concluded clinical separability does not strongly correlate with downstream RL performance. They did not account for training stability of the CDE autoencoder, and trained for a fixed epoch length. In fact, when the CDE training is properly stabilized, both the MSE loss and the acuity correlation loss govern the same underlying latent quality, leading to representations that yield high RL policy returns and capture clinically meaningful trajectories. Appendix~\ref{app:WIS_comparison_acuity_alignment} further verifies this conclusion by comparing the RL policies trained on CDE representations with and without acuity correlation regularization.

\subsection{Stabilization Method Comparison}\label{sec:exp_stabilization}
We evaluate three stabilization methods using our early stopping criteria, with all methods using $\lambda=1$ for acuity correlation. Each method was trained for 100 epochs with optimal hyperparameters from Table~\ref{tab:stabilization_hparams}. For each method, we compute \textbf{flatness metrics} mentioned in Section~\ref{sec:stabilizing_method} on the loss curve, respectively \textbf{plateau length} and \textbf{mean absolute slope on plateau ($\mathbf{S_1}$)}. Then we freeze each encoder at its \textit{optimal and stable} checkpoint, train a downstream dBCQ RL policy using identical setting: policy learning rate \(=1\times10^{-5}\), dBCQ action elimination threshold \(\tau_{\text{BC}}=0.3\), and epochs \(=2\times10^{5}\), and get their mean evaluated WIS returns on validation trajectories.

\begin{table}[t]
\centering
\begin{tabular}{>{\centering\arraybackslash} p{1.8cm} 
                >{\centering\arraybackslash} p{2cm} 
                >{\centering\arraybackslash} p{2.3cm} 
                >{\centering\arraybackslash} p{0.8cm}}
\toprule
\makecell[c]{\textbf{Method}} 
  & \makecell[c]{\textbf{Plateau}\\\textbf{Length}} 
  & \makecell[c]{$\mathbf{S_{1}}$} 
  & \makecell[c]{$\mathbf{\overline{WIS}}$} \\
\midrule
Baseline  & $42\pm 8.6$ & $0.0028\pm 0.0005$ &  $0.9195$\\
Gradient clipping & $45.8\pm 4.5$ & $0.0031\pm 0.0021$ & $0.6547$ \\
Implicit Adams solver & $46.5\pm 5.7$ & $0.0027\pm 0.0006$ & $0.9206$ \\
Stiffness regularization & $43\pm 9.4$ & $0.0030\pm 0.0008$ &  $0.9189$\\
\bottomrule
\end{tabular}
\caption{Flatness metrics and downstream RL WIS return for each stabilization method (under best hyperparameters, across random seeds).}
\label{tab:stabilization_results}
\end{table}
As shown in Table~\ref{tab:stabilization_results}, the implicit Adam solver produces the longest plateau, lowest mean absolute slope, and achieves the highest mean RL return ($\overline{WIS} = 0.9206$). Both implicit Adam solver and stiffness regularization yield comparable RL performance w.r.t baseline, while gradient clipping weakened both the downstream RL policy performance and flatness of training curve. 

\begin{figure}[t]
    \centering
    \includegraphics[width=\columnwidth]{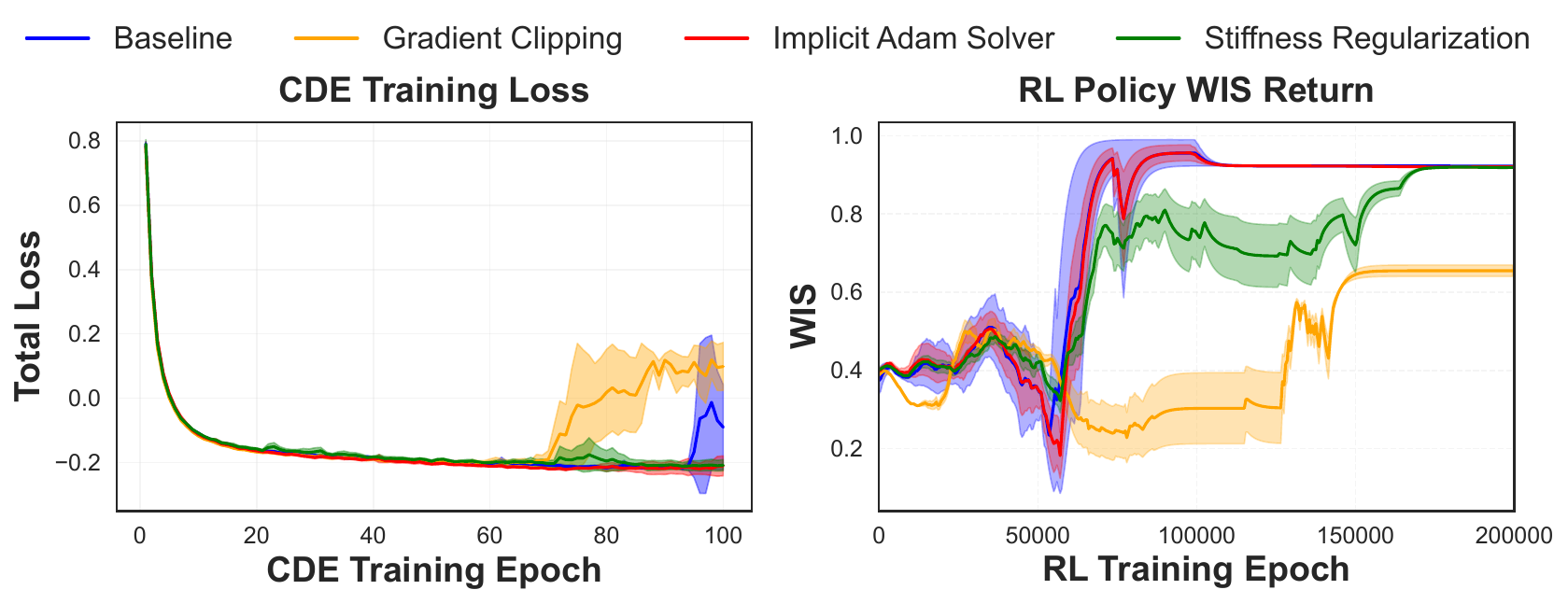}
    \caption{The CDE autoencoder training loss ($mean \pm std$) and WIS return ($mean \pm std$) for downstream RL on validation trajectories across random seeds for each stabilization method.}
    \label{fig:stabilization_methods_comparison}
\end{figure}

Figure~\ref{fig:stabilization_methods_comparison} demonstrates that when training RL on the CDE representations from baseline and implicit Adam solver, the downstream RL policy converges fast to an optimal WIS return; The stiffness regularized representation converges slower, but achieves a comparable WIS return finally. The gradient clipping representation, however, converges to a suboptimal WIS return, and has the most erratic CDE training curvature.
As a result, gradient clipping is not suitable for this clinical application, as it may overly dampen the dynamics of the learned vector field; Implicit Adam solver is the best method to balance training stability and representation quality.
\section{Conclusion}
Our work establishes practical guidelines for training Neural CDEs in clinical RL, addressing critical gaps in representation training stability and clinical alignment that were overlooked in prior works. We demonstrate that stabilizing CDE autoencoder training through early stopping and stabilization methodologies is essential for obtaining clinical meaningful and robust state representations, and could also significantly improve downstream RL policy performance. Our proposed multi-criteria early stopping strategy enables downstream RL policies to achieve superior performance (WIS $> 0.9$ compared to $\sim 0.775$ in \cite{killian20a}), while stabilization methods like implicit Adam solver provide additional robustness by extending stable training plateaus. 

We reconcile the controversy around acuity scores by showing they improve both RL policy returns and clinical interpretability when paired with stabilized CDE.

Our correlation analysis reveals a deeper insight that the state representation quality and acuity correlation are fundamentally linked by a shared latent structure. This shared latent space property explains why our stabilized training simultaneously achieves both high policy returns and clinical interpretability.

\clearpage
\bibliographystyle{named}
\bibliography{ijcai25}
\clearpage
\appendix
\section*{Appendix}
\section{Details about Patient Cohort}\label{app:patient_cohort}
We follow the approach from \cite{Komorowski2018Nature} to process the sepsis cohort from the MIMIC-III v1.4 critical care database. The original extracted data contains 48 variables including demograpics, elixhauser status, laboratory results, vital signs, fluids and vasopressors, and fluid balance. There are missing and irregulaly sampled data in original MIMIC-III dataset, following the method by \cite{Komorowski2018Nature},
19,418 adult sepsis patients were selected from the MIMIC-III database by applying the criteria: \begin{itemize}
    \item Patients are aged 18 or older;
    \item An acute increase in the Sequential Organ Failure Assessment (SOFA) score of 2 or more;
    \item Exclude admissions where treatment was withdrawn or mortality was not documented.  
\end{itemize}
Then the data was filled using a time-limited approach based on clinically relevant periods for each variable, then further imputed using nearest neighbour algorithm.
We then select 33 features (Table~\ref{tab:time_varying_features}) that are most relevant to sepsis treatment. 
\begin{table*}[t]
    \centering
    \renewcommand{\arraystretch}{1.3}
    \begin{tabular}{llll}
    \toprule
    \textbf{Feature 1--8} & \textbf{Feature 9--16} & \textbf{Feature 17--24} & \textbf{Feature 25--33} \\
    \midrule
    Glasgow Coma Scale & Potassium & White Blood Cells  & PaO\textsubscript{2}/FiO\textsubscript{2} \\
    Heart Rate         & Sodium    & Platelets  & Bicarbonate (HCO\textsubscript{3}) \\
    Systolic BP        & Chloride  & PTT  & SpO\textsubscript{2} \\
    Diastolic BP       & Glucose   & PT   & BUN \\
    Mean BP            & INR       & Arterial pH  & Creatinine \\
    Respiratory Rate   & Magnesium & Lactate  & SGOT \\
    Body Temp (°C)     & Calcium & PaO\textsubscript{2}  & SGPT \\
    FiO\textsubscript{2}& Hemoglobin & PaCO\textsubscript{2}   & Bilirubin \\
                       &     &  & Base Excess \\
    \bottomrule
    \end{tabular}
    \caption{List of 33 time-varying continuous physiological features used for state representation training.}
    \label{tab:time_varying_features}
\end{table*}

For reinforcement learning in this task, actions are defined as combinations of intravenous (IV) fluids and vasopressors, discretized into 25 clinically meaningful bins based on percentile ranges. 
As shown in Table~\ref{tab:rl_action_space}, these bins span from no administration to higher dose quartiles, forming a $5\times 5$ action space, comprising 25 distinct treatment actions.
Different types of vasopressors are converted to noradrenaline-equivalent, and the unit is mcg/kg/min. IV fluids are corrected for tonicity and converted to a standard unit.
\begin{table*}[t]
    \centering
    \renewcommand{\arraystretch}{1.2}
    \begin{tabular}{cccccc}
    \toprule
    \textbf{Action Number} & \textbf{0} & \textbf{1} & \textbf{2} & \textbf{3} & \textbf{4}\\
    \midrule
    Vasopressors & $0.00$ & $(0.00, 0.08]$ & $(0.08, 0.22]$ & $(0.22, 0.45]$ & $>0.45$\\
    IV fluids & $0.00$ & $(0.00, 50.00]$ & $(50.00, 180.00]$ & $(180.00, 530.00]$ & $>530.00$\\
    \bottomrule
    \end{tabular}
    \caption{Discretized dosage bins for vasopressors and intravenous (IV) fluids used to define the 25-action reinforcement learning space}
    \label{tab:rl_action_space}
\end{table*}
\section{Acuity score alignment}\label{app:acuity_score}
Patient acuity scores are used to measure the severity of a patient's condition, and are crucial for clinical decision-making. To constrain the learning of state representations, we extract three acuity scores from the full patient observations from each 4 hour time step (in MIMIC‑III dataset):
\begin{itemize}
    \item \textbf{SOFA (Sequential Organ Failure Assessment)} \cite{Vincent1996SOFA}: Assesses dysfunction across respiratory, coagulation, liver, cardiovascular, CNS, and renal systems. Scores range from 0 (normal) to 24 (severe failure).
    \item \textbf{SAPS-II (Simplified Acute Physiology Score)} \cite{LeGall1993SAPSII}: Predicts ICU mortality using 17 physiological measurements. Scores range from 0 to 163, with higher values indicating greater mortality risk.
    \item \textbf{OASIS (Oxford Acute Severity of Illness Score)} \cite{Johnson2013OASIS}: Estimates mortality risk from 10 clinical variables. Scores range from 10 to 83, where higher scores correspond to worse prognosis.
\end{itemize}
These scores are used to regularize the CDE latent space by the Pearson correlation between the state representation and the acuity socres. Equation~\ref{eq:acuity_correlation_loss} defines the acuity correlation regularized loss, where we choose the hyperparameter $\lambda$ to be the same for all three acuity scores, yet these could be chosen independently of one another and yield a loss function :
\begin{equation}
\begin{aligned}
    &\mathcal{L}_{\text{total}}(o_{t}, \hat{o}_{t}) = \mathcal{L}_{\text{MSE}}(o_{t}, \hat{o}_{t}) \\
    - &\left(\lambda_1 \cdot \rho_{\text{SOFA}}(\hat{s}_t)
    + \lambda_2 \cdot \rho_{\text{OASIS}}(\hat{s}_t)
    + \lambda_3 \cdot \rho_{\text{SAPSII}}(\hat{s}_t) \right)
    \label{eq:acuity_correlation_loss}
\end{aligned}
\end{equation}
\subsection{Acuity score heatmaps}\label{app:acuity_score_heatmap_app}
In section~\ref{sec:exp_acuity_alignment}, we have compared the SOFA score heatmap of \textit{optimal and stable} and \textit{overtrained and unstable} CDE representations. Here we provide the heatmaps of other acuity scores, including OASIS and SAPS-II. The heatmaps are generated by projecting the latent features of validation trajectories into a lower dimensional space using PCA, and coloring the points by their respective acuity scores.
\begin{figure}[t]
    \centering
    \includegraphics[width=\linewidth]{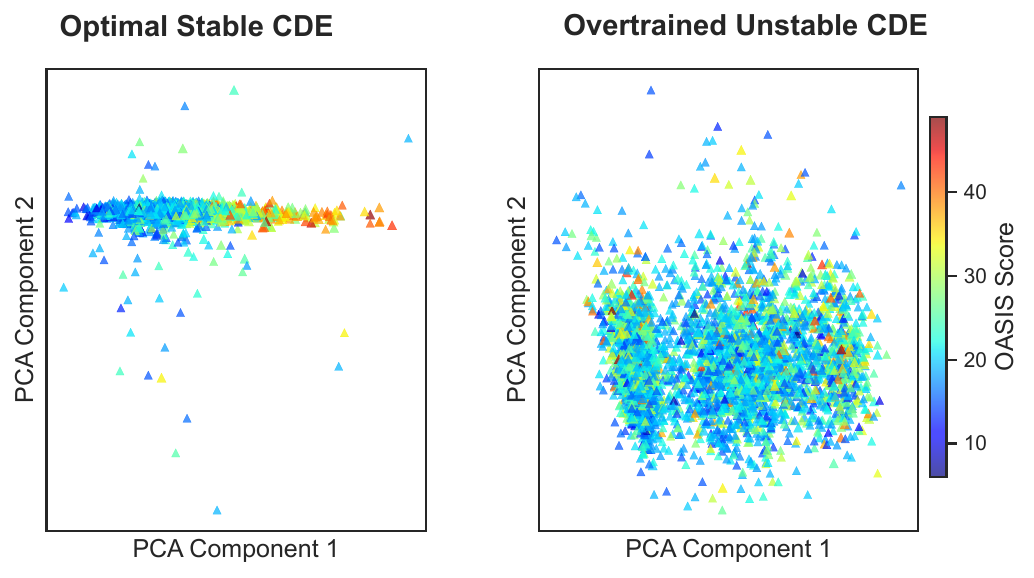}
    \caption{Heatmap of OASIS score distribution at \textit{optimal and stable} and \textit{overtrained and unstable} CDE representations on validation dataset}
    \label{fig:OASIS_heatmap}
\end{figure}
\begin{figure}[t]
    \centering
    \includegraphics[width=\linewidth]{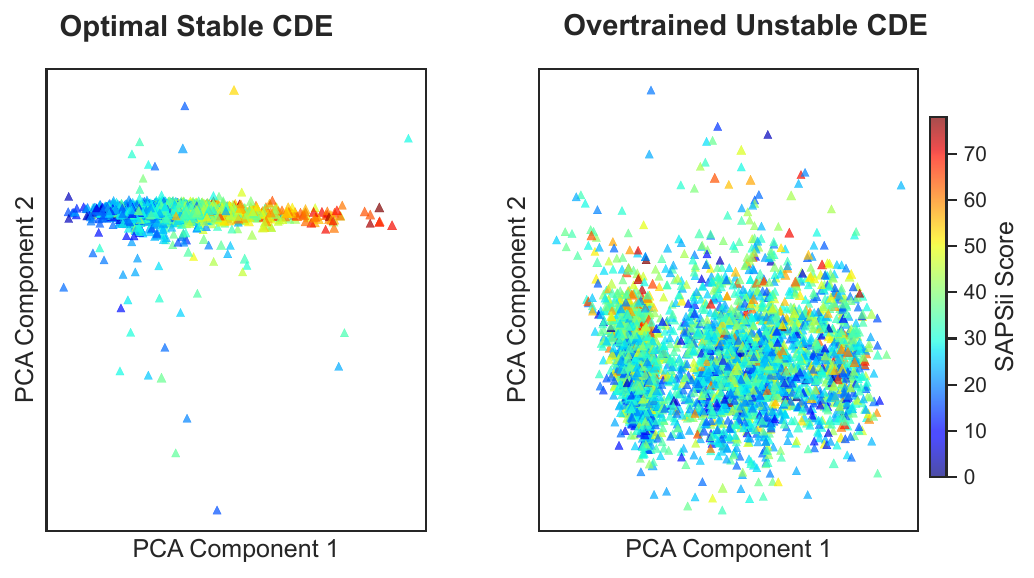}
    \caption{Heatmap of SAPS-II score distribution at \textit{optimal and stable} and \textit{overtrained and unstable} CDE representations on validation dataset}
    \label{fig:SAPSII_heatmap}
\end{figure}
The hyperparameters used for generating these heatmaps are the same as those used in Section~\ref{sec:exp_stopping_criteria}, hidden size = 64, learning rate = $2\times 10^{-4}$, $\lambda = 1$, $\epsilon_1 = 0.1$, $p=30$, $\epsilon_2 = 0.02$, $\rho_{\text{threshold}}=0.7$.

As shown in Figure~\ref{fig:OASIS_heatmap} and Figure~\ref{fig:SAPSII_heatmap}, the \textit{optimal and stable} CDE representation shows a clear clustering of patients with similar OASIS and SAPS-II scores, while the \textit{overtrained and unstable} CDE representation exhibits a more scattered and random distribution, indicating a loss of clinical alignment. This further supports our finding that early stopping is crucial for maintaining acuity score alignment in CDE representations.
\subsection{Correlation between MSE loss and Acuity Correlation loss}\label{app:mse_acuity_correlation}
\begin{figure}[t]
    \centering
    \includegraphics[width=\linewidth]{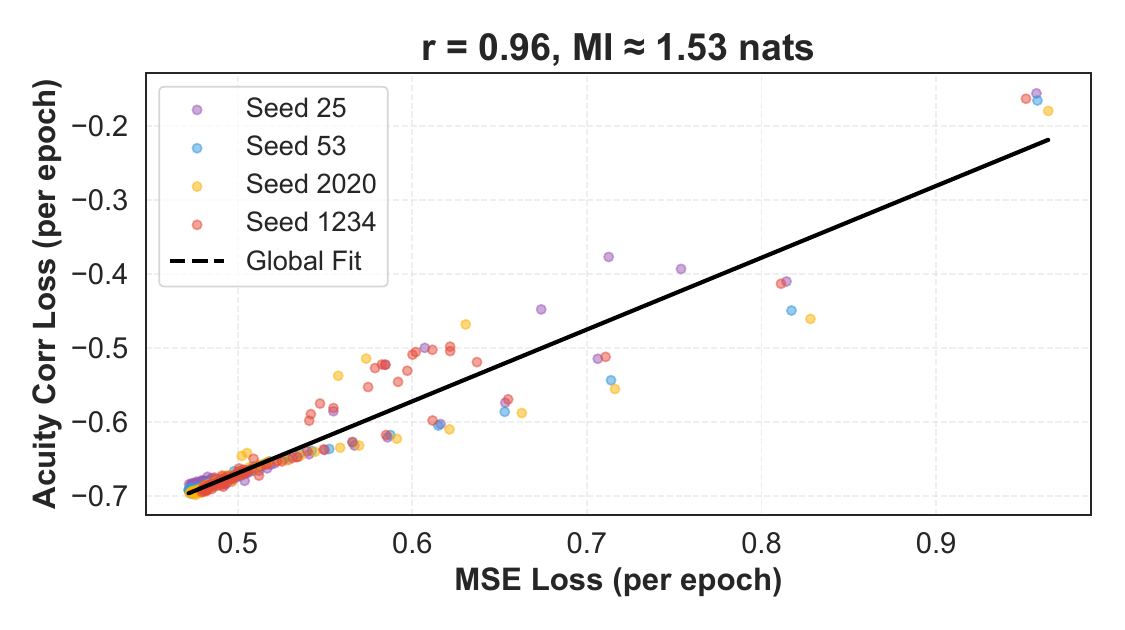}
    \caption{Scatter plot of MSE loss ($\mathcal{L}_{\text{MSE}}$) and Acuity Correlation loss ($\mathcal{L}_{\text{corr}}$) during CDE autoencoder training, across random seeds. The x-axis represents the MSE loss, and the y-axis represents the acuity correlation loss. Each scatter point represents the an epoch during CDE autoencoder training. The blackline represents the linear regression fit to the data. The Pearson correlation coefficient is $0.9578\pm 0.0127$ with a significant p-value $<10^{-45}$; The mutual information is $1.5296 \pm 0.1847$ (nats).}
    \label{fig:mse_acuity_correlation_scatter}
\end{figure}

In Section~\ref{sec:exp_acuity_alignment}, we have mentioned that the MSE loss and acuity correlation loss are consistent during CDE autoencoder training. Here we provide the mutual information plots between MSE loss and acuity correlation loss, which further supports our finding that the MSE loss and acuity correlation loss are capturing the same underlying latent quality. The mutual information is calculated using the empirical distribution of the losses over training epochs. As shown in Figure~\ref{fig:mse_acuity_correlation_scatter}, the scatter plot of MSE loss and acuity correlation loss shows a strong linear relationship, with a high Pearson correlation coefficient with significant p-value; The high mutual information of $1.5296 \pm 0.1847$ nats indicates that the MSE loss and acuity correlation loss are capturing the same underlying latent quality, which further verifies our finding that early stopping is crucial for maintaining acuity score alignment in CDE representations.
\subsection{RL policy comparison on CDE representations with / without acuity alignment}\label{app:WIS_comparison_acuity_alignment}
In this section we compare the downstream RL performance on CDE representations with and without acuity alignment. We use the CDE representations trained with ($\lambda=1$) and without ($\lambda=0$) acuity correlation loss, respectively, and train a dBCQ RL policy on each representation. Hyperparameters in Table~\ref{tab:stop_criteria} are tuned respectively to get an \textit{optimal and stable} checkpoint for both settings. 

\begin{figure}[t]
    \centering
    \includegraphics[width=\linewidth]{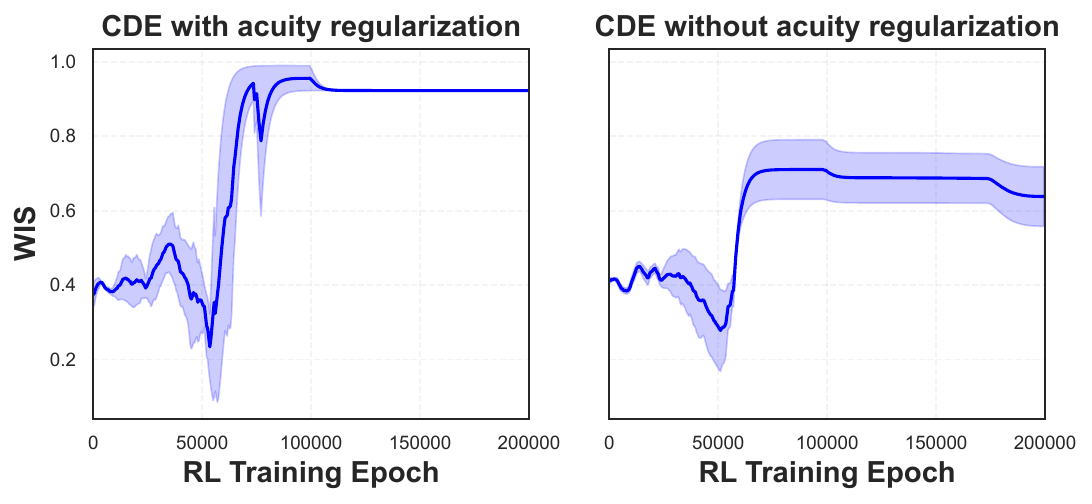}
    \caption{Comparison of WIS returns of dBCQ RL policies trained on CDE representations with and without acuity alignment, the left plot denotes the CDE autoencoder trained with $\lambda=1$ on its best hyperparameters, the right plot denotes the CDE autoencoder trained with $\lambda=0$ on its best hyperparameters. The x-axis represents the RL training epochs, and the y-axis represents the WIS returns.}
    \label{fig:WIS_comparison_acuity_alignment}
\end{figure}

The results are shown in Figure~\ref{fig:WIS_comparison_acuity_alignment}, where we can see that the dBCQ RL policy trained on CDE representations with acuity alignment achieves higher WIS returns $0.9195$ than the one trained on CDE representations without acuity alignment $0.6381$. This further supports our finding that acuity correlation 
\section{Early Stopping Criteria}\label{app:early_stopping_criteria}
Table~\ref{tab:stop_criteria} summarizes the best hyperparameters for the early stopping criteria in Section~\ref{sec:cde_stopping_criteria}. The hyperparameters are tuned through grid search, and the best performing configuration is selected based on the downstream RL policy performance measured by WIS return.
\begin{table*}[ht]
    \centering
    \begin{tabular}{p{3cm} p{7.5cm} p{2.5cm} p{3.4cm}}
      \toprule
      \textbf{Criterion} & \textbf{Metric} & \textbf{Hyperparameter} & \textbf{Best Hyperparameter} \\
      \midrule
      Low Validation Loss & Final validation loss should reach a minimal value during training & \(\epsilon_{1}\) & $0.1$\\
  
      Plateau & Number of consecutive epochs with small relative loss change & \makecell[l]{\(p\)\\\(\epsilon_{2}\)} & \makecell[l]{$30$\\$0.02$}\\
  
      Acuity Correlation & Mean Pearson correlation between learned features and acuity scores & \(\rho_{\text{threshold}}\) & $0.7$\\
      \bottomrule
    \end{tabular}
   \caption{Criteria for selecting optimal stopping epoch in CDE autoencoder training with hyperparameters to be tuned.}
   \label{tab:stop_criteria}
  \end{table*}
\section{Stabilization methods}\label{app:stabilization_methods}
\begin{figure*}[t]
    \centering
    \includegraphics[width=\textwidth]{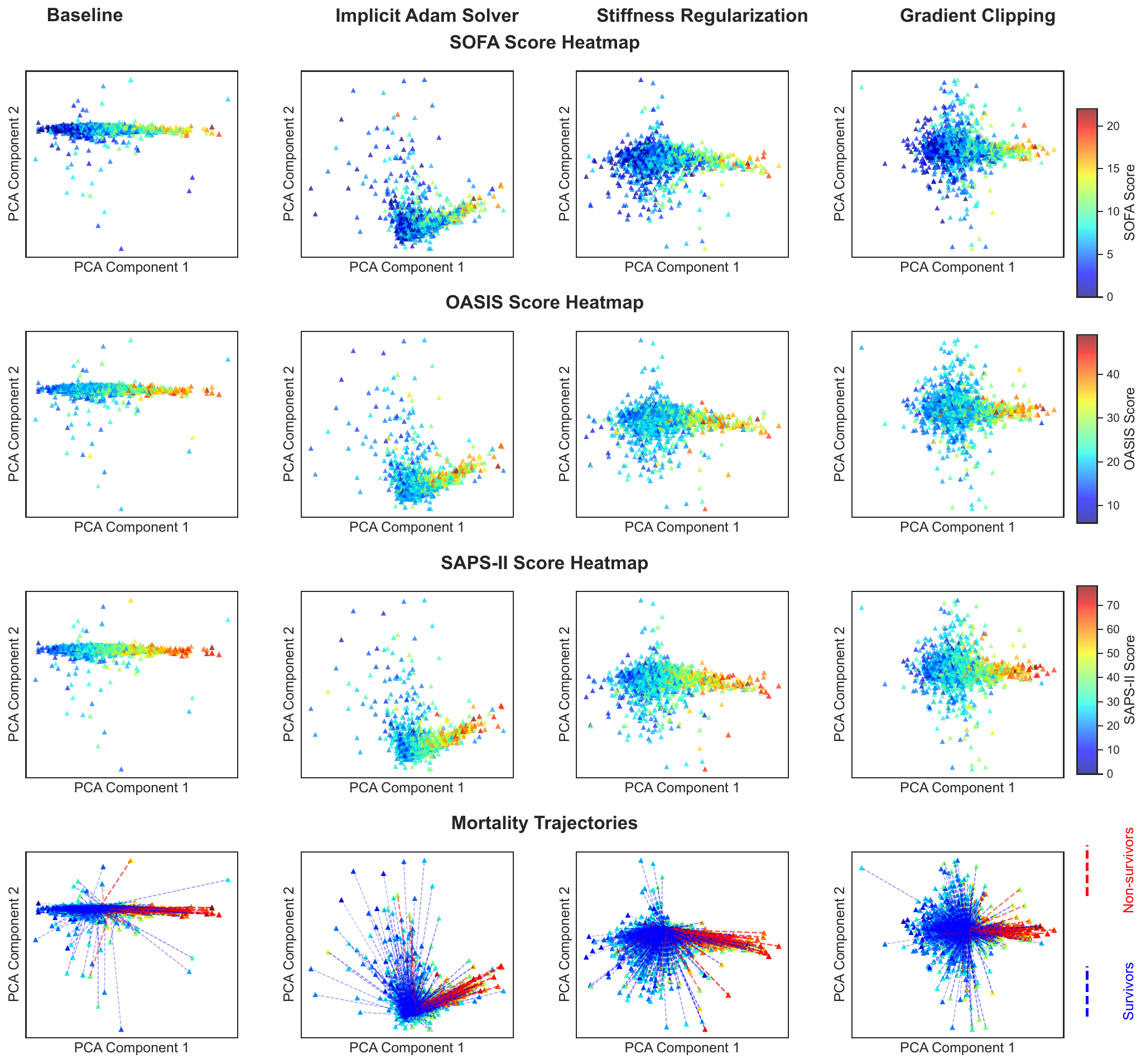}
    \caption{SOFA, OASIS, SAPS-II score distribution and mortality trajectories of CDE autoencoders for each stabilization method on validation dataset. It can be shown that following the early stopping criteria, all methods achieve a clear clustering of patients with similar acuity scores, and the non-survivor trajectories are well separated from survivor trajectories.}
    \label{fig:cde_acuity_comparison_stabilization_methods}
\end{figure*}

In this section, we provide a detailed description of the stabilization methods used in Section~\ref{sec:exp_stabilization} to stabilize CDE autoencoder training. We evaluate three methods: implicit Adams solver, gradient clipping, and stiffness regularization. Each method is designed to improve the stability of the CDE autoencoder training process.
Following the early stopping criteria, we evaluate the effectiveness of these methods by measuring the RL policy performance on the CDE representations, and also show the acuity alignment of each stabilization method on validation dataset in Figure~\ref{fig:cde_acuity_comparison_stabilization_methods}.
\subsection{Implicit Adams solver}\label{app:implicit_adams_solver}
When training the Neural CDE latent state $h(t)$ evolving under observation $o_t$ as in Equation~\ref{eq:CDEformula}, implicit Adams methods advance $h(t)$ using future values of the target function in the integration step.
An $s$ step Adams-Moulton update rule from step $t_{n}$ to $t_{n+1}$ is defined as follows \cite{debrouwer2023anamnesicneuraldifferentialequations}:
\begin{align}
    h(n+1) = h(n) + \Delta t\sum_{j=0}^{s}\beta_j\cdot f_{\theta}(h(n+1-j), o_{t_{n+1-j}})
\label{eq:implicit_adams_moulton}
\end{align}
where $\Delta t = t_{n+1} - t_n$ and $\beta_j$ are fixed coefficients. 
Note from Equation~\ref{eq:implicit_adams_moulton} that both sides include the term $f_{\theta}(h(n+1-j), o_{t_{n+1-j}})$, so a nonlinear equation needs to be solved at each step. 
Here we list orders 0, 1, 2 and 4 for Adams-Moulton:
\begin{align}
    h(n) =& h(n-1) + \Delta t \cdot f_{\theta}(h(n), o_{t_{n}}); \\
    h(n+1) &= h(n) + \Delta t\cdot \Big(\frac{1}{2}f_{\theta}(h(n+1), o_{t_{n+1}})\notag\\
           &+\frac{1}{2}f_{\theta}(h(n), o_{t_{n}})\Big); \\
    h(n+2) &= h(n+1) + \Delta t\cdot\Big(\frac{5}{12}f_{\theta}(h(n+2), o_{t_{n+2}})\notag \\
    + &\frac{8}{12}f_{\theta}(h(n+1), o_{t_{n+1}}) - \frac{1}{12}f_{\theta}(h(n), o_{t_{n}})\Big);\\
    h(n+4) &= h(n+3) + \Delta t\cdot\Big(\frac{251}{270}f_{\theta}(h(n+4), o_{t_{n+4}}) \notag\\
    + \frac{646}{720}&f_{\theta}(h(n+3), o_{t_{n+3}})- \frac{264}{720}f_{\theta}(h(n+2), o_{t_{n+2}}) \notag \\
    + \frac{106}{720}&f_{\theta}(h(n+1), o_{t_{n+1}}) - \frac{19}{720}f_{\theta}(h(n), o_{t_{n}})\Big)
\end{align}
In our experiments in Section~\ref{sec:exp_stabilization}, we used the implicit Adams-Moulton method of order 4, which is a 4th-order implicit method. 
We do a grid search for \textit{step size} $\Delta t\in \{\frac{1}{8}, \frac{1}{4}, \frac{1}{2}\}$, and the best performance is achieved with $\Delta t=\frac{1}{8}$, resulting in a mean WIS return of $0.9206$ on the downstream RL policy.

As shown in Figure~\ref{fig:stabilization_methods_comparison} and Table~\ref{tab:stabilization_results}, the implicit solver not only makes the curvature of the loss function smoother, makes the plateau longer and more stable, but also achieves the best performance in downstream RL policy.
This indicates that implicit solver method balances the trade-off between smoothness and extracting sharp transits in the latent space (e.g, sudden deteriorations in septic patients), which is crucial for ICU time-series data.
\subsection{Gradient clipping}\label{app:gradient_clipping}
Norm-based gradient clipping is a technique that limits the magnitude of the gradient vector during backpropagation. Given a parameter vector $\theta$ with gradient $g = \nabla_\theta L$ at a training step, the Euclidean norm $\|g\|_{2}$ is computed, then a threshold $\tau>0$ is chosen, and the gradient is rescaled if its norm exceeds $\tau$. Formally, the clipped gradient $g_{\text{clip}}$ is defined by:
\begin{equation}
    g_{\text{clip}} = \begin{cases}
        g, & \text{if }\|g\|_{2}\le\tau\\
        \tau\frac{g}{\|g\|_{2}} & \text{if }\|g\|_{2}>\tau
    \end{cases}
\label{eq:clipped_gradient}
\end{equation}
Mathematically, gradient clipping can be seen as a form of regularization on the optimization trajectory. It does not alter the loss function or objective explicitly, but it modifies the optimization dynamics to avoid excessively large parameter jumps. 

When training Neural CDE model, as introduced in Equation~\ref{eq:CDEformula}, one backpropagates through the numerical solver to compute the gradients of the loss function with respect to the model parameters. If $f_{\theta}$ has large Lipschitz constant or the integration time is long, the backward gradients can suffer from gradient explosion.
Previous works \cite{NEURIPS2022Gradient,COELHO2024} explicitly cite gradient clipping as an effective method for controlling gradient explosion and oscillatory in continuous‐time models.

In section~\ref{sec:stabilizing_method}, we apply gradient clipping with a grid search on the threshold $\tau\in\{0.5, 1.0, 1.5\}$, and as a result, when $\tau=1.0$, the CDE autoencoder achieves the best performance in downstream RL policy. With a mean WIS return of $0.6547$, it is lower than the baseline CDE autoencoder without stabilization, which achieves a mean WIS return of $0.9195$. An explanation for this could be that in irregular ICU time-series, patient trajectories are sparese and short, and gradient clipping uniformly dampens high-magnitude updates, however, some large gradients might be clinically meaningful.
Also, in ICU trajectories, there are sudden deteriorations in septic patients, which causes large updates to $\theta$, while gradient clipping will suppress that update, leading to a less informative representation.
\subsection{Stiffness Regularization}\label{app:l2_regularization}
This regularization technique leverages stiffness indicators to shape the training dynamics of Neural CDEs. Recall from Section~\ref{sec:CDE} that the encoder is defined as :
\begin{equation}
    \partial h(t) = f_{\theta}(h(t))\partial o_t
\end{equation}
Assume that we solve it over time interval $[t_0, t_1]$, and reconstruct the downstream target $\hat{o}_{t_1}=\phi(h(t_1))$. Let $\mathcal{L}_{\text{total}}$ in Equation~\ref{eq:loss_func_with_acuity} be the reconstruction loss, $\{t_j\}_{j=1}^{N}$
be the solver time steps. Specifically, for each time step $t_j$, the stiffness score $\mathcal{S}_j$ is apprixmated via the real parts of the eigenvalues of the local Jacobian:
\begin{equation}
    \mathcal{S}_j = \max\left\{|\Re (\lambda_{i})| : \lambda_{i}\in\text{eig}\left(\frac{\partial f_{\theta}(h(t_{j}))}{\partial h(t)}\right)\right\}
\end{equation}
where\begin{itemize}
    \item $\frac{\partial f_{\theta}(h(t_{j}))}{\partial h(t)}\in \mathbb{R}^{d\times d}$ denotes the Jacobian of the vector field w.r.t. the hidden state at time $t_j$.
    \item eig($\cdot$) denotes the set of eigenvalues.
    \item $\Re(\lambda_i)$ denotes the real part of $\lambda_i$. 
\end{itemize}
Then the total loss becomes:\begin{equation}\label{eq:regularized_objective}
    \mathcal{L}_{\text{reg}} = \mathcal{L}_{\text{total}} + \lambda_{\text{reg}} \cdot \underbrace{\sum_{j=1}^{N} S_j}_{\text{Stiffness Regularization}}
\end{equation}
This objective encourages the encoder's neural vector field $f_{\theta}$ to not only fit the data but also generate dynamics that are less stiff, leading to smoother trajectories in the latent space. 
The hyperparameter $\lambda_\text{reg}$ controls the strength of this regularization, which balances the tradeoff between fitting the data and minimizing the solver complexity.
As mentioned in Table~\ref{tab:stabilization_hparams}, we do a grid search on $\lambda_\text{reg}\in\{0.005, 0.01, 0.015\}$, and find that $\lambda_\text{reg}=0.01$ achieves the best performance in downstream RL policy, with a mean WIS return of $0.9189$.
Figure~\ref{fig:stabilization_methods_comparison} shows that the stiffness regularization stabilizes the CDE training and lead to a longer, smoother plateau, which is beneficial for picking the optimal stopping epoch. And its downstream RL policy reaches a comparable optimal WIS return. 

In summary, stiffness regularzation is a promising method for stabilizing CDE autoencoder training for learning MIMIC-III data representation, as it encourages the model to learn smoother dynamics, which is crucial for capturing the underlying patterns in irregular ICU time-series data. It also helps to extend the plateau of the training loss, making it easier to find the optimal stopping epoch.

\section{Limitation and Future Work}\label{app:limitations}
When evaluating the state representations by RL policy, the bias-variance tradeoff in Weighted Importance Sampling (WIS) is tricky. Particulary, for sepsis treatment, the variance of WIS is highly sensitive to the behaviour policy threshold $\tau_{\text{BC}}$. As $\tau_{\text{BC}}$ decreases, the variance of WIS gets higher.
Like the prior works, our work does not discuss potential pitfalls of WIS variance with respect to $\tau_{\text{BC}}$ and how to mitigate it. Future work should analyze the impact of $\tau_{\text{BC}}$, and explore strategies such as doubly robust estimators to reduce WIS variance.

In the future, we plan to extend our work to more stabilization methods including Gaussian noise injection \cite{alain2014regularizedautoencoderslearndata} and dropout in vector field \cite{NEURIPS2018RegularizedAutoencoders}. To evaluate the state representations, other SOTA offline RL methods such as Fitted Q-Iteration, Conservative Offline Model Based Policy Optimization should be considered. There is still big potential to extend acuity alignment to other time-series medical tasks including septic shock prediction \cite{Giannini2019SepsisML}, ICU stroke recovery \cite{Choo2022MLStrokeRehab}, and post-surgical complication prediction \cite{Hassan2023AISurgery}.

\end{document}